\documentclass[letterpaper, 10 pt, conference]{ieeeconf}  

\IEEEoverridecommandlockouts                              

\usepackage{cite}
\usepackage{amsmath,amssymb,amsfonts}
\usepackage{mathtools}
\usepackage{multirow}
\usepackage{float}
\usepackage{algorithmic}
\usepackage{graphicx}
\usepackage{textcomp}
\usepackage{xcolor}
\usepackage{pifont}
\newcommand{\cmark}{\ding{51}}%
\newcommand{\xmark}{\ding{55}}%
\usepackage[hidelinks]{hyperref}
\hypersetup{
  colorlinks,
  allcolors=.,
  urlcolor=blue,
}

\DeclareMathOperator*{\E}{\mathbb{E}}

\hypersetup{
  colorlinks,
  allcolors=.,
  urlcolor=blue,
}

\title{\scriptsize This preprint has been accepted to IEEE RA-L for publication. Copyright may be transferred without notice, after which this version may no longer be accessible. \\ \LARGE \bf
Learning to Predict Navigational Patterns from Partial Observations
}

\author{Robin Karlsson$^{1*}$, Alexander Carballo$^{2}$, Francisco Lepe-Salazar$^{3}$, \\ Keisuke Fujii$^{1}$, Kento Ohtani$^{1}$, Kazuya Takeda$^{1,4}$
\thanks{$^{1}$Graduate School of Informatics, Nagoya University, Aichi, Japan.}
\thanks{$^{2}$Department of Electrical, Electronic and Computer Engineering, Gifu University, Gifu, Japan.}
\thanks{$^{3}$Ludolab, Colima, M\'exico}
\thanks{$^{4}$TIER IV, Tokyo, Japan}
\thanks{*Corresponding author: karlsson.robin@g.sp.m.is.nagoya-u.ac.jp}
}

\begin{document}

\maketitle
\thispagestyle{plain}
\pagestyle{plain}

\begin{abstract}
Human beings cooperatively navigate rule-constrained environments by adhering to mutually known navigational patterns, which may be represented as directional pathways or road lanes.
Inferring these navigational patterns from incompletely observed environments is required for intelligent mobile robots operating in unmapped locations. However, algorithmically defining these navigational patterns is nontrivial.
This paper presents the first self-supervised learning (SSL) method for learning to infer navigational patterns in real-world environments from partial observations only.
We explain how geometric data augmentation, predictive world modeling, and an information-theoretic regularizer enable our model to predict an unbiased local directional soft lane probability (DSLP) field in the limit of infinite data.
We demonstrate how to infer global navigational patterns by fitting a maximum likelihood graph to the DSLP field.
Experiments show that our SSL model outperforms two SOTA supervised lane graph prediction models on the nuScenes dataset.
We propose our SSL method as a scalable and interpretable continual learning paradigm for navigation by perception. Code is available at \url{https://github.com/robin-karlsson0/dslp}.

\end{abstract}

\section{Introduction}

Mobile robots perform tasks that involve traversing an environment. 
To navigate rule-constrained structured environments robots are required to correctly perceive and interpret the environment. This problem is called scene understanding.
Navigational patterns, or directional pathways, are a core component of understanding how to traverse structured environments~\cite{krause2013robot_navigation}.
In particular, efficient and safe multi-agent navigation depends on each agent following mutually known navigational patterns. The patterns can be defined by explicit rules or be derived from social conventions and emergent behavior.
However, learning to infer navigational patterns for complex environments based on observable features is difficult due to regional variation and noise including varying or missing surface markings, geometries, and materials.

Current methods for spatial navigation can be categorized into mapping- and learning-based approaches. The mapping approach~\cite{sheif2016hdmaps} avoids the problem of automatized understanding of environments by encoding human knowledge in the form of lane maps and localizing the system within these maps.
Creating a priori navigation maps is a conceptually simple, interpretable, and predictable way to safely navigate environments.
In practice, this approach is difficult to scale

\begin{figure}[H]
\centering
\includegraphics[width=0.48\textwidth]{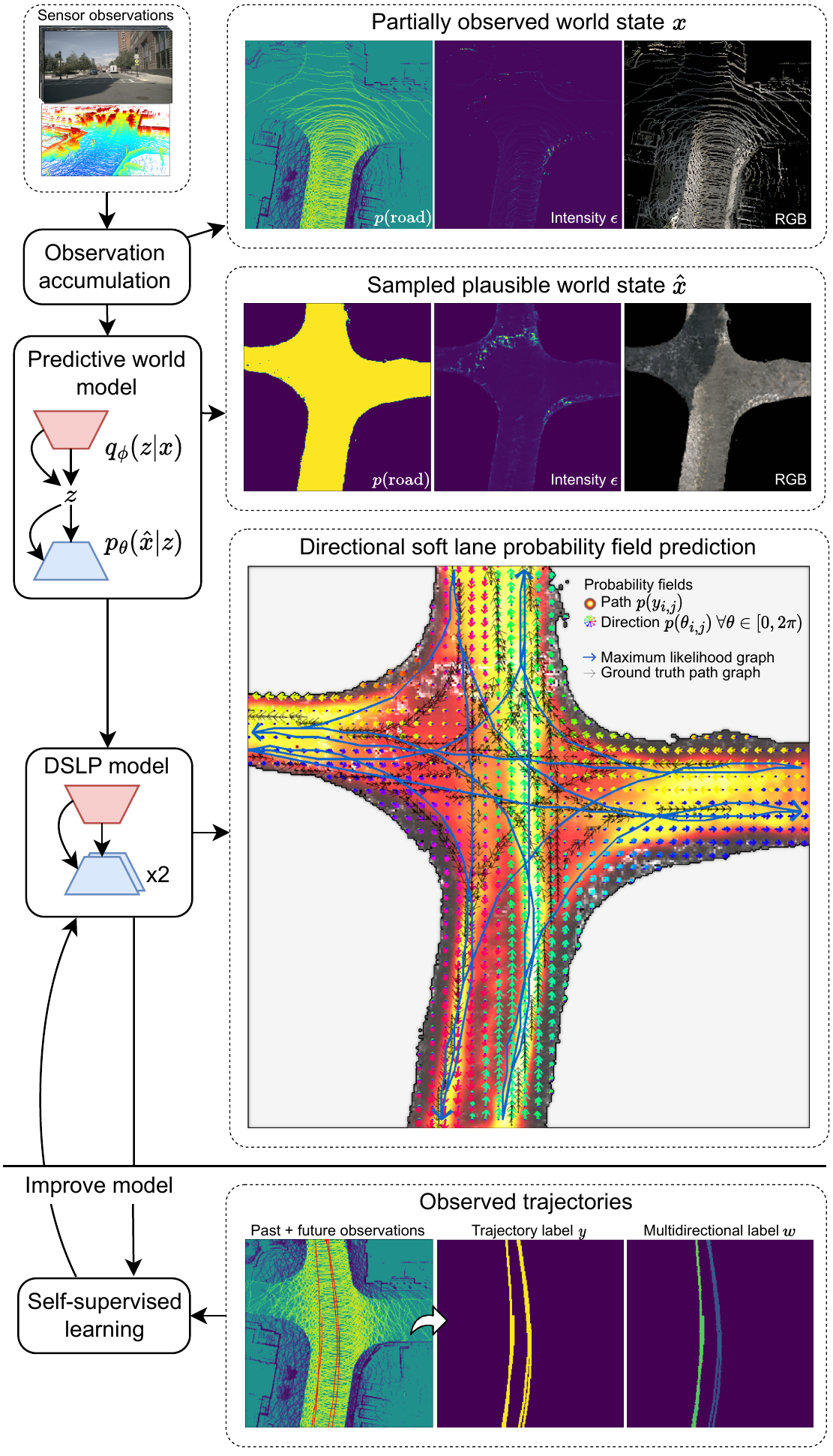}
\vspace{-8mm}
\caption{The method accumulates sensor observations into a common metric vector space representing the partially observed world state $x$. A predictive world model samples a set of diverse plausible complete world states $\hat{x}$. The directional soft lane probability (DSLP) model predicts two probability fields; the agent traversal probability $p(y_{i,j})$ and a multimodal directional probability distribution $p(\theta_{i,j})$ for each point $(i,j)$. A fitted maximum likelihood graph corresponds to global navigational patterns. The DSLP model can learn navigational patterns from observed trajectories representing only a subset of all plausible trajectories.}
\label{fig:front_figure}
\vspace{-3mm}
\end{figure}

\noindent up, as map creation, maintenance, and verification are costly in terms of human labor, typically limiting application to small predetermined environments. Additionally, dynamic navigational behavior like correctly avoiding parked cars or debris cannot be a priori encoded in static maps.

The learning approach involves training a model to infer navigational patterns based on environmental context. Some methods learn implicit patterns as part of accomplishing the primary task~\cite{bojarski2016e2e_self_driving, amin2018variational_end2end, bansal2018}. Other methods learn explicit patterns but require ground truth lane maps for training~\cite{zurn2021lanegraphnet, can2021stsu}.
Methods learning from observational data alone are promising scalable solutions to infer navigational patterns, as driving data can be obtained at a low cost.
However, the real-world performance of existing methods is fragile and unpredictable in complex environments and lacks interpretability.

The human visual system comprises two subsystems~\cite{gibson1979visual_system, milner2008visual_systems, han2022modeling_ventral_dorsal}.
The vision-for-perception system located in the ventral stream processes information in a slow, top-down manner to create perceptual representations from ambiguous or incomplete visual input by leveraging visual and semantic memory~\cite{milner2008visual_systems}. These representations support conscious mental processes such as recognition, visual thought, and planning.
The vision-for-action system located in the dorsal stream processes information in a real-time, bottom-up manner to perceive the entire environment and infer behaviorally-relevant visual affordances, including cues for spatial navigation~\cite{milner2008visual_systems, milner2012visual_processing_conscious}.
%

In this paper we present a self-supervised method for learning to infer navigational patterns from real-world partial observations as required for traversing unmapped real-world environments.
Our approach is inspired by the biological dorsal visual pathway~\cite{milner2008visual_systems} and endows artificial intelligent agents with a functionally similar self-improving system that learns to infer visual affordances for spatial navigation~\cite{sheth2016visual_pathways_sampling_space}.

The model learns general contextual environment features that explain observed trajectories, and can thus infer navigational patterns for newly encountered environments. Learning from observed trajectories means learning from only a subset of all plausible trajectories. We propose an information-theoretic regularizer to overcome the problem of false negative traversal observations resulting from partial observations.
Our model combines complementary aspects of mapping- and learning-based approaches. It also produces an interpretable representation akin to maps. Lastly, this model improves with additional experience akin to continual learning~\cite{lesort2020cont_learn_robot} while avoiding catastrophic forgetting by retaining a replay buffer of past experiences~\cite{minh2015dqn}.

We identify the navigational pattern prediction problem based on static environmental context as a sub-problem of the general dynamic agent behavior prediction problem. The main difference is that we do not consider the influence of dynamic objects such as parked cars and red traffic lights, or predict the movement of particular agents.
While both problems can be solved through the same framework, we choose to remove dynamic object information from the input representation in order to objectively compare performance against ground truth lane graph methods.

While we perform experiments in a real-world urban road environment our method is applicable in any general structured environment.



The contributions of our paper are three-fold:
\begin{itemize}
    \item A self-supervised approach for learning to predict unbiased traversability probability maps from real-world partial positive-only observations using a principled hyperparameter-free information-theoretic regularizer.
    \item Experimentally show that our method improves with additional observations and achieves better performance than recent state-of-the-art (SOTA) supervised methods.
    \item Experimentally verify that leveraging a predictive world model~\cite{karlsson2023pred_wm} and geometric data augmentation~\cite{karlsson2020dsla} improves real-world performance.
\end{itemize}

The rest of the paper is organized as follows. Sec. II reviews the SOTA and contrasts it with our work. Sec. III explains how partial observations are transformed into complete world states used as model input. Sec. IV explains our method and model implementation. Sec. V. explains the experiment setup. Sec. VI. present experimental results. Sec. VII concludes the paper
by discussing limitations and future improvements to our method.

\section{Related work}

\noindent \textbf{Path prediction.} Recent works present methods to predict multimodal paths for specific actors.
Salzmann~et al.~\cite{salzmann2019online_path_generation} and Baumann~et~al.~\cite{baumann2018ego_path} trains a convolutional neural network (CNN) on bird's-eye-view (BEV) environment representations to predict a dense map representing valid ego-vehicle paths using a weighted dense classification error and future ego-vehicle trajectories. Barnes~et~al.\cite{barnes2017ws_path_proposal} trains a CNN on perspective images with self-supervised labels generated from driving data. Ort~et~al.~\cite{ort2020maplite} fuses high-level navigational guidance from a coarse map with path generation reflecting the observed environment. Casas~et~al.~\cite{casas2021mp3} optimizes a model to predict an environment map and possible paths for the ego-agent based on images and point clouds using a ground truth lane map as supervision.
Prez-Higueras~et~al.~\cite{prez-higueras2018point_to_point_path} trains a CNN model to predict a multimodal path affordance map between any two points to be used as a prior for an $\text{RRT}^*$ path planner~\cite{karaman2011RRT_star}. 
Kitani~et~al.~\cite{kitani2012activity_forcasting} trains a Hidden Parameter Markov Decision Process (HiPMDP) model
using inverse reinforcement learning and observation data. Ratliff~et~al.~\cite{ratliff2009learch} presents an imitation learning approach that maps input features to a cost map based on example paths.
Our approach expands on prior works by learning to  predict all plausible navigational patterns in the environment independently of observed agents without depending on ground truth maps for supervision.

\noindent \textbf{Lane graph and map prediction.}
Homayounfar~et~al.~\cite{homayounfar2018hran} trains a recurrent neural network (RNN) model to predict polylines as road lanes in highway road scenes using ground truth lane maps. An extension~\cite{homayounfar2019dagmapper} introduces forking and merging lane topologies.
Guo~et~al.~\cite{guo2020genlanenet} predicts 3D road lanes from perspective images using ground truth annotations.
Z\"urn~et~al.~\cite{zurn2021lanegraphnet} trains a Graph-RCNN model to predict lane anchors and edges using images and point clouds with ground truth lane map supervision.
Can~et~al.~\cite{can2021stsu} trains a transformer model to detect lane segments from images and subsequently connected into lane graphs.
Zhang~et~al.~\cite{zhang2021hierarchical_road_topology_learning} trains a three-stage network using ground truth map supervision to predict a dense lane map and subsequently predict keypoints used to generate the graph.
Mi~et~al.~\cite{mi2021hdmapgen} presents a hierarchical coarse-to-fine approach to train an attention graph neural network to generate road lane graphs.
Karlsson~et~al.~\cite{karlsson2020dsla} presents a self-supervised method to train a directional soft lane affordance (DSLA) map from single trajectories. A follow-up work~\cite{karlsson2021gdsla} shows how to generate discrete road lane graphs by searching for connected paths in the DSLA map using the $\text{A}^*$ algorithm. 
Our method is a scalable approach to predict lane graphs from partial observations without requiring ground truth lane map annotations and yet achieve better performance than supervised baselines~\cite{zurn2021lanegraphnet, can2021stsu}. This work extends~\cite{karlsson2020dsla,karlsson2021gdsla} by introducing a principled regularizer, a sampling-based maximum likelihood graph generation method, and demonstrates the approach on real-world data.

Another line of works consider the problem of predicting a structured semantic representation of the environment akin to human-annotated HD maps~\cite{sheif2016hdmaps} from sensor observations and ground truth maps. 
Li~et~al.~\cite{li2022hdmapnet} trains a multimodal network to predict dense maps subsequently post-processed into vectorized representations of map elements.
An extension~\cite{liu2023vectormapnet} directly predicts vectorized map elements.
Liao~et~al.~\cite{liaomaptr} presents a transformer model trained end-to-end to predict vectorized map elements from camera images.
Shin~et~al.~\cite{shin2023instagram} presents an attention graph neural network approach.
Ort~\cite{ort2022maplite2} presents a model-based approach to fit parametric map elements according to observations and prior map information.
Our approach is complementary as it provides explicit navigational patterns based on an environment representation.

\noindent \textbf{End-to-end learning for autonomous vehicles.}
Originally proposed by Pomerleau~\cite{pomerleau1988alvinn} and more recently repopularized by Bojarski~et~al.~\cite{bojarski2016e2e_self_driving}, the end-to-end learning paradigm aims to learn a driving model or policy mapping perception to control by optimizing for an extrinsic goodness objective.
Imitation learning approaches~\cite{bojarski2016e2e_self_driving, amin2018variational_end2end, bansal2018} learn a policy that results in similar behavior as expert examples.
Reinforcement learning (RL) approaches~\cite{kendall2019end2end_deeprl} optimize a policy to maximize an extrinsically defined reward such as time-to-human-override.
Recently, approaches learning an explicit predictive world model~\cite{henaff2019model_predictive_policy_learning, chen2021world_on_rails} show that robust policies can be learned from expert observation only.
Our method to learn explicit agent-agnostic navigational patterns is an alternative approach to enhance explainability of end-to-end learning, or incorporate an end-to-end learning aspect into the conventional modularized mobile robotics system~\cite{krause2013robot_navigation}.






\section{Plausible world state input generation}
\label{sec:plausible_world_state_input_generation}

Here we describe the pre-processing method shown in Fig.~\ref{fig:front_figure}. Sensor observations are accumulated into partially observed world states $x$, which in turn are transformed into plausible world states $\hat{x}$. The proposed model
uses $\hat{x}$ as input.

\subsection{Partial world state representation}
\label{sec:partial_world_state_representation}

We generate partial world states based on accumulated sensor observations following the method described in prior work~\cite{karlsson2023pred_wm}. The method shares similarities with a hierarchical biological model of human representation and processing of visual information~\cite{mar1982human_hierarchical_vision_repr}.
The agent is initialized within an unknown metric vector space. Sensor observations are projected onto this common vector space at discrete timesteps. Semantic information is inferred from images using a pretrained semantic segmentation model and appended to coincident 3D points to form semantic point clouds.
Past semantic point clouds are integrated with new observations by scan matching using the ICP algorithm~\cite{vizzo2022kiss-icp} and SLAM~\cite{smith1986slam} for loop closure. The accumulated semantic point cloud is reduced to a five-layered 2D probabilistic BEV representation $x \in \mathbb{R}^{I \times J \times C}$ with dimension $I \times J$ elements, and $C$ denoting the number of semantic information channels. In this work, $C$ consists of five channels representing the semantic attributes of a spatial point $(i,j)$; we represent road probability $p(\text{road})$ by a beta distribution, lidar reflection intensity $\epsilon$ as a scalar value, and visual appearance by RGB values.

Dynamic objects are detected by a pretrained object detection model and represented by 3D bounding boxes. Trajectory observations are generated by temporally tracking detected objects. 
Dynamic objects are considered ``moving'' if motion is observed or ``static'' otherwise. This classification allows filtering away observations associated with moving dynamic objects while keeping observations of static dynamic objects for training, as they may influence how other agents navigate the environment such as swerving out of the lane to avoid a parked car. The static dynamic objects can be removed at inference time to provide an agent-agnostic prediction of navigational patterns akin to a lane map.

\begin{figure}
\centering
\includegraphics[width=0.48\textwidth]{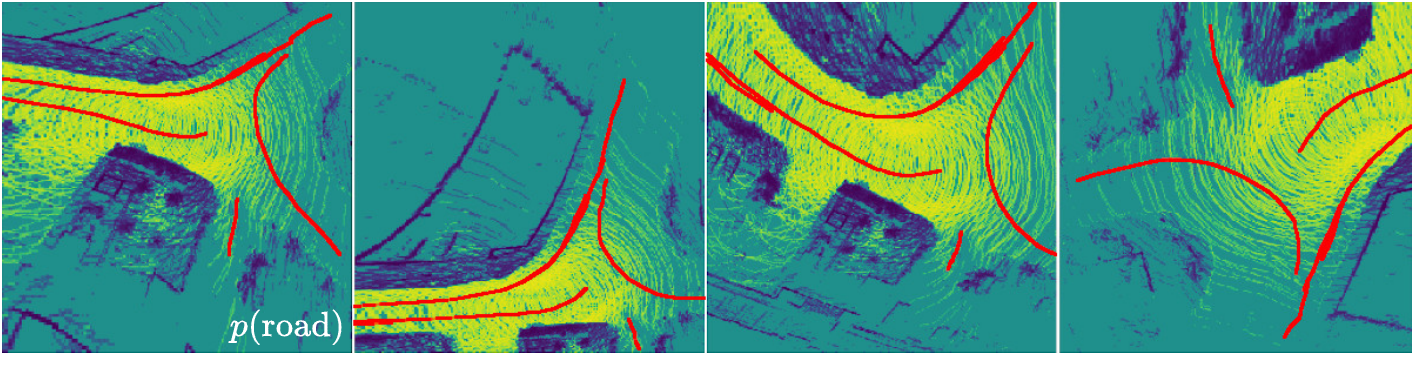}
\vspace{-8mm}
\caption{Geometric data augmentation generates diverse sample variations from a single real sample. Spatial information (dense maps) and observed trajectories (red lines) are transformed by the same function.}
\label{fig:data_aug}
\vspace{-5mm}
\end{figure}

We leverage geometric data augmentation~\cite{karlsson2020dsla} to improve model generalization performance by learning geometric invariance. Each sample is augmented by random rotation and translation, and a polynomial warping function is applied to the dense maps and observed trajectories
\begin{equation}
    a_0(\xi^{\prime})^2 + a_1 (\xi^\prime) + a_2 = \xi
\label{eq:data_aug}
\end{equation}
\noindent where $\xi$ is a substitute for spatial coordinates $i$ and $j$, and $\xi^\prime$ denotes warped coordinates. We create dense warp maps by using the inverse function of (\ref{eq:data_aug}) to map each warped coordinate $\xi^\prime$ to an original coordinate $\xi$. The coefficients $a_0$, $a_1$, and $a_2$ are derived by satisfying boundary conditions~\cite{karlsson2020dsla}. Fig.~\ref{fig:data_aug} shows visual examples of a sample augmentation.

\subsection{Predictive world model}
The predictive world model~\cite{karlsson2023pred_wm} samples diverse and plausible complete world states $\hat{x}$ conditioned on partially observed world states $x$ as exemplified in Fig.~\ref{fig:front_figure}. The world model is functionally similar to the biological ventral cortical pathway as the model disambiguates the partially observed environment by leveraging past experience~\cite{milner2008visual_systems}.
The world model is computationally conceptualized as an arbitrary conditioning generative model and implemented by the recent SOTA hierarchical VAE (HVAE) model VDVAE~\cite{child2021vdvae} with the encoder module replaced by a posterior matching encoder~\cite{karlsson2023pred_wm}. The HVAE models the joint distribution of observable variables $p(r, \epsilon, R, G, B)$ factorized as the conditional distribution
\begin{equation}
    p(r, \epsilon, R, G, B) = p(R|G,B,r) p(G|B,r) p(B|r) p(\epsilon|r) p(r)
\end{equation}
\noindent using hierarchical latent variables $z$. Here $r$ and $\epsilon$ denote road and lidar reflection intensity, and RGB are image color channels. The latent variable prior 
$p(z)$ and posterior $q(z|x)$ distributions are factorized as
\begin{align}
 p(z) &= p(z_1 | z_2) \ldots p(z_{K-1} | z_K) p(z_K) \label{eq:hvae_prior} \\
 q(z|x) &= q(z_1 | z_2, x) \ldots q(z_{K-1} | z_K, x) q(z_K | x)
\end{align}
\noindent with random variables $z$ modeled by normal distributions. 

\begin{figure}
\centering
\includegraphics[width=0.40\textwidth]{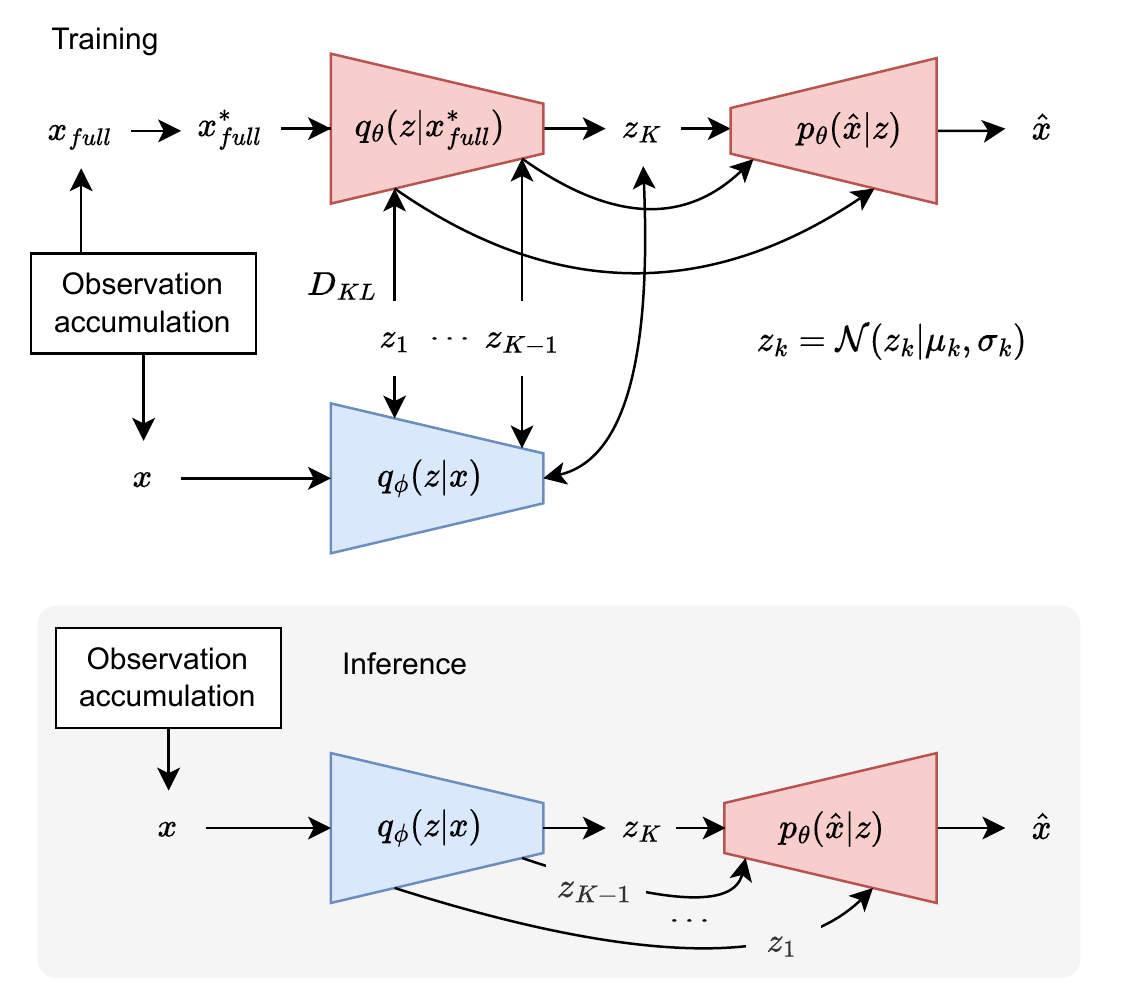}
\vspace{-3mm}
\caption{(Top) The predictive world model is trained to reconstruct pseudo ground-truth world states $x^\ast_{full}$ and simultaneously optimize a secondary encoder $q_\phi(z|x)$ to predict a similar hierarchical latent distribution $z$ as the primary encoder $q_\theta(z|x^\ast_{full})$ from partially observed world states $x$. (Bottom) The trained model samples a hierarchical latent distribution $z$ from $x$ to generate complete plausible world states $\hat{x}$.}
\label{fig:pred_wm}
\vspace{-5mm}
\end{figure}

The world model learns to approximate the prior and posterior distributions by the parameterized models $q_\theta(z|x)$ and $p_\theta(x|z)$ using variational inference~\cite{kingma2013vae} and trained using self-supervised learning to predict future observations from present observations akin to the predictive coding problem~\cite{marino2019pred_coding_vae_bio}. Note that the vanilla HVAE cannot learn to generate diverse complete representations from partially observed representations only. We follow the posterior matching optimization method visualized in Fig.~\ref{fig:pred_wm} and presented in prior work~\cite{karlsson2023pred_wm} to overcome this limitation. The method trains a regular HVAE using pseudo ground-truth world states $x^\ast_{full}$, and a secondary encoder $q_\phi(z|x)$ to predict a similar hierarchical latent distribution $z = \{z_1, \ldots, z_K\}$ as the primary encoder $q_\theta(z|x^\ast_{full})$ from $x$.

We generate pseudo ground-truth world states $x^\ast_{full}$ using a sequential process starting from the intermediate world state $x_{full}$ consisting of past and future observations as explained in prior work~\cite{karlsson2023pred_wm}.
The regular HVAE model is trained by maximizing the hierarchical ELBO over $x_{full}$~\cite{child2021vdvae, karlsson2023pred_wm}.
%
%
%
%
The second encoder is optimized by minimizing

\begin{equation}
\begin{gathered}
 D_{KL}(q_\theta(z|x^\ast_{full}) || q_\phi(z|x)) = \\
 \sum^K_{k=1} \E_{q(z_{>k}|x)} \left[ D_{KL}( q_\theta(z_k | z_{>k}, x^\ast_{full}) || q_\phi(z_k | z_{>k}, x)) \right].
\end{gathered}
\end{equation}

At inference time the model uses the partially observed encoder to generate a latent distribution $q_\phi(z|x)$ that can be decoded by $p_\theta(\hat{x}|z)$ into a completely observed plausible world state $\hat{x}$ similar to a pseudo ground-truth world state $x^\ast_{full}$ without the need to observe the future.



\section{Directional soft lane probability model}
\label{sec:dslp_model}

\begin{figure}
\centering
\includegraphics[width=0.40\textwidth]{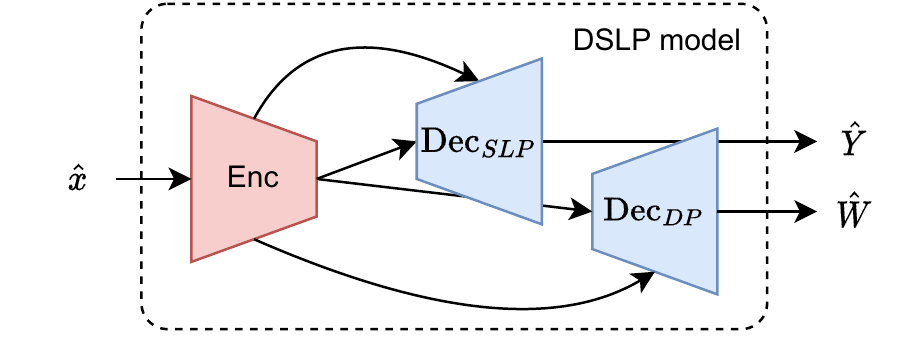}
\vspace{-3mm}
\caption{The Directional Soft Lane Probability (DSLP) model uses a dual decoder U-Net~\cite{ronneberger2015unet} model to transform a plausible world state $\hat{x}$ into a soft lane probability (SLP) map $\hat{Y}$ and directional probability (DP) tensor $\hat{W}$.}
\label{fig:dslp_model}
\vspace{-5mm}
\end{figure}

Here we present a method to train a model to predict unbiased probability maps of local directional traversability. The model input is the plausible world state $\hat{x}$ described in Sec.~\ref{sec:plausible_world_state_input_generation}. We also present a method for inferring global navigational patterns from the local probability maps. See Fig.~\ref{fig:front_figure} and Fig.~\ref{fig:output_viz} for output visualizations.

The model is implemented by a U-Net neural network~\cite{ronneberger2015unet} with a single encoder and two decoders as illustrated in Fig.~\ref{fig:dslp_model}. The first decoder outputs a probability map $Y \in \mathbb{R}^{I \times J}$ representing soft lane probabilities for elements in a grid map of size $I \times J$. The second decoder outputs a map of categorical distributions $W \in \mathbb{R}^{M \times I \times J}$ representing $M$ direction interval probabilities for each location $(i,j)$. The methods for optimizing both probabilistic outputs are explained below.

\subsection{Soft Lane Probability (SLP) Modeling}

The likelihood of each environment location $(i,j)$ being traversed by an unspecified agent is modeled by the predicted probability value $\hat{y}_{i,j} \in \hat{Y}$ and is called soft lane probability (SLP). Learning to predict an unbiased $\hat{Y}$ from partial observations is nontrivial, as the self-supervised learning signal contains false negative traversal observations (i.e. lacking an observed trajectory where traversals are probable). We formalize the problem as follows. Ideally we want to learn a distribution $q(y)$ that approximates the true distribution $p(y)$. However, optimizing $q(y)$ according to the learning signal results in learning the distribution of partially observed samples $\tilde{p}(y)$. A principled solution is to use a regularizer to decrease bias and make $q(y)$ better match $p(y)$.

In this paper we present a semi-supervised objective that enables learning an unbiased probabilistic prediction of traversability based on an information-theoretic regularizer derived from balancing the information contribution from positive and negative partial observations in $Y$.

In information theory, the entropy $H(y)$ of a distribution $p(y)$ is considered a quantity that measures information content. The cross-entropy
\begin{equation}
    H(p,q) \triangleq - \sum_{k=1}^K p(y=k) \: \log ( q(y=k) )
\end{equation}
\noindent measures the information overhead to compress a sample $y \sim p(y)$ using a code based on $q(y)$~\cite{murphy2022pml1}.

Each partial observation $Y$ contains two distinct groups of traversal information; a set of true positives representing certain information, and a set of true and false negatives representing uncertain information. The contributed information of the set of positive and negative observations are
\begin{equation}
    H(Y_{pos} \subseteq Y, \hat{Y}) = - \sum_{i,j \in Y_{pos}} y_{i,j} \: log ( \hat{y}_{i,j} )
\label{eq:info_contrib_pos}
\end{equation}
\begin{equation}
    H(Y_{neg} \subseteq Y, \hat{Y}) = - \sum_{i,j \in Y_{neg}} (1-y_{i,j}) \: log ( 1 - \hat{y}_{i,j} ).
\label{eq:info_contrib_neg}
\end{equation}
\noindent We devise a regularizer based on balancing the information contribution provided by (\ref{eq:info_contrib_pos}) and (\ref{eq:info_contrib_neg}) according to the ratio of observations
\begin{equation}
    \alpha_{IB} = |Y_{pos}| \; / \; (|Y_{pos}| + |Y_{neg}|)
\label{eq:alpha_sla}
\end{equation}
where $|Y_{*}|$ denotes the number of positive and negative observed elements $(i,j)$. The balanced information contribution $H^*(Y|\hat{Y})$ is obtained by linearly interpolating the information contributions according to the ratio of observations
\begin{equation}
    H^*(Y|\hat{Y}) = \alpha_{IB} \: H(Y_{neg}|\hat{Y}) + (1-\alpha_{IB}) \: H(H_{pos}|Y).
\end{equation}

Linear interpolation is a monotonic function that balances the information contributions while preserving the total information quantity
\begin{equation}
    0 \le H^*(Y|\hat{Y}) \le \max(H(Y_{pos}|\hat{Y}), H(Y_{neg}|\hat{Y})).
\end{equation}

We formulate the problem specific optimization objective $\mathcal{L}_{SLP}$ as the mean balanced information contribution
\begin{equation}
\begin{gathered}
  \mathcal{L}_{SLP} = - \frac{1}{|Y|} \sum_{i,j \in Y} [ \alpha_{IB} (1-y_{i,j}) log (1-\hat{y}_{i.j}) \\ + (1-\alpha_{IB}) y_{i,j} log (\hat{y}_{i,j}) ]
\end{gathered}
\label{eq:slp_objective}
\end{equation}
\noindent where $\hat{y}_{i,j}$ and $y_{i,j}$ is the predicted and observed soft lane probability for the element located at $i,j$. $|Y|$ denotes the number of traversable elements. The information contribution ratio $\alpha_{IB}$ provides the optimal interpolation between positive and negative traversal observations.


One can view (\ref{eq:slp_objective}) as the cross entropy objective with an additional dynamic regularizer between positive and negative observations. Experiments show that the balanced information contribution cross-entropy objective (\ref{eq:slp_objective}) performs better than finetuning a static hyperparameter weighting~\cite{karlsson2020dsla}, and allows learning probabilistic predictions despite occasional abnormal observations unlike the barrier loss objective~\cite{karlsson2021gdsla}.

The negative log likelihood $\text{NLL}_{\text{SLP}}$ of an observed sample $y$ according to a model prediction $\hat{y}$ based on modeling $p(y|\hat{y})$ as a Bernoulli distribution is

\begin{equation}
    \text{NLL}_{\text{SLP}} = - \sum_{i,j \in Y} \left[ y_{i,j} \: log (\hat{y}_{i,j}) + (1-y_{i,j}) \: log (1 - \hat{y}_{i,j}) \right].
\label{eq:nll_slp}
\end{equation}



\subsection{Directional Probability (DP) Modeling}

The likelihood of local traversal directionality at each location $(i,j)$ is modeled by the predicted vector $\hat{w}_{i,j}$ called directional probability (DP). The $\hat{w}_{i,j}$ models a categorical probability distribution representing the direction interval $\theta \in [0, 2 \pi)$ by $M$ uniformly spaced intervals
\begin{equation}
    w_{i,j} = (p(\theta \in [0, \frac{2\pi}{M})), \ldots , p(\theta \in \left[\frac{(M-1) 2 \pi}{M}, 2 \pi \right)) )^T.
\end{equation}

The learning signal is created by encoding observed trajectories into $w_{i,j}$ as a discrete von Mises distribution. In the case of multiple overlapping trajectories the individual distributions are superimposed and renormalized. Learning to match distributions improve multimodal prediction compared with learning to predict single values by maximum likelihood estimation~\cite{karlsson2020dsla}.

The optimization objective $\mathcal{L}_{DP}$ is formulated as learning to predict the directional distribution by minimizing the mean KL divergence between predicted $\hat{w}_{i,j}$ and observed $w_{i,j}$ directionality over all elements $w_{i,j} \in W$

\begin{equation}
    \mathcal{L}_{DP} = \frac{1}{|W|} \sum_{i,j \in W} D_{KL}(w_{i,j} || \hat{w}_{i,j}).
\label{eq:da_objective}
\end{equation}

Note that the learning signal used to optimize the DP objective (\ref{eq:da_objective}) lacks false negatives and therefore does not require regularization like the SLP objective (\ref{eq:slp_objective}).

The negative log likelihood $\text{NLL}_{\text{DP}}$ of an observed sample $w_{i,j}$ according to a model prediction $\hat{w}_{i,j}$ based on modeling $p(w|\hat{w})$ as a categorical distribution is 

\begin{equation}
    \text{NLL}_{\text{DP}} = - \sum_{i,j \in Y} \sum_{m=1}^M w^{(m)}_{i,j} \: log ( \hat{w}^{(m)}_{i,j} ).
\label{eq:nll_dp}
\end{equation}


\subsection{Maximum likelihood lane graph}
\label{seq:maximum_likelihood_path_graph}

\begin{figure}
\centering
\includegraphics[width=0.48\textwidth]{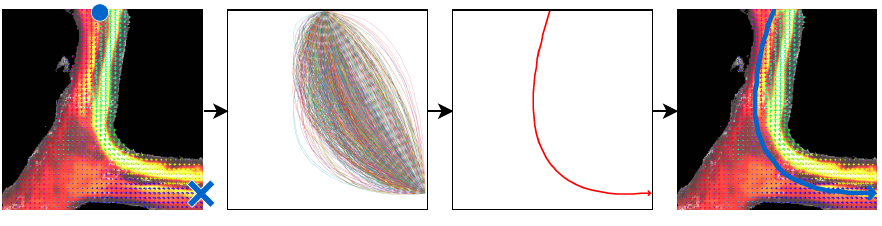}
\vspace{-9mm}
\caption{The maximum likelihood graph is generated by connecting entry~($\bullet$) and exit~($\times$) points by the most probable of many sampled paths given the predicted DSLP field.}
\label{fig:maximum_likelihood_graph}
\vspace{-6mm}
\end{figure}

Evaluating the goodness of local navigational patterns using the predicted DSLP field is straightforward. To also evaluate the usefulness of the predicted DSLP field for inferring global navigational patterns, we present a sampling-based method to generate a maximum likelihood road lane graph fitted to the predicted DSLP field. The graph generation process is illustrated in Fig.~\ref{fig:maximum_likelihood_graph}.

First, we infer entry and exit points at the edges of the predicted DSLP field. A non-maximum suppression (NMS) operation is performed on the SLP field $\hat{Y}$ to find the most likely path centers. Each point is designated as an entry and/or exit point according to the predicted DP field $\hat{W}$. Additional entry and exit points are inferred from directional field regions which are coherent but lack a NMS point.

Secondly, we incrementally build a graph by searching for valid connecting paths between all entrance and exit points by a sampling-based approach. A set of second-degree polynomial spline paths is generated between an entry and exit pair by randomly sampling a valid spline control point $(i,j)^*$ from a normal distribution with rejection sampling.
The likelihood of each sampled path is evaluated using the location and directionality of $M$ equidistant points along the path given the predicted DSLP field using (\ref{eq:nll_slp}) and (\ref{eq:nll_dp}). The path with the lowest total NLL is selected as the best path. Repeating this process results in a set of most likely paths representing the maximum likelihood graph. A post-processing operation removes undesired edges between neighboring lanes (i.e. u-turns) using a simple distance threshold heuristic.
Representing navigational patterns by splines is a useful inductive bias, as agents tend to navigate structured environments in a continuous and smooth manner.

\section{Experiments}

\begin{figure}[t]
\centering
\includegraphics[width=0.48\textwidth, trim={0 1.5cm 0 0}]{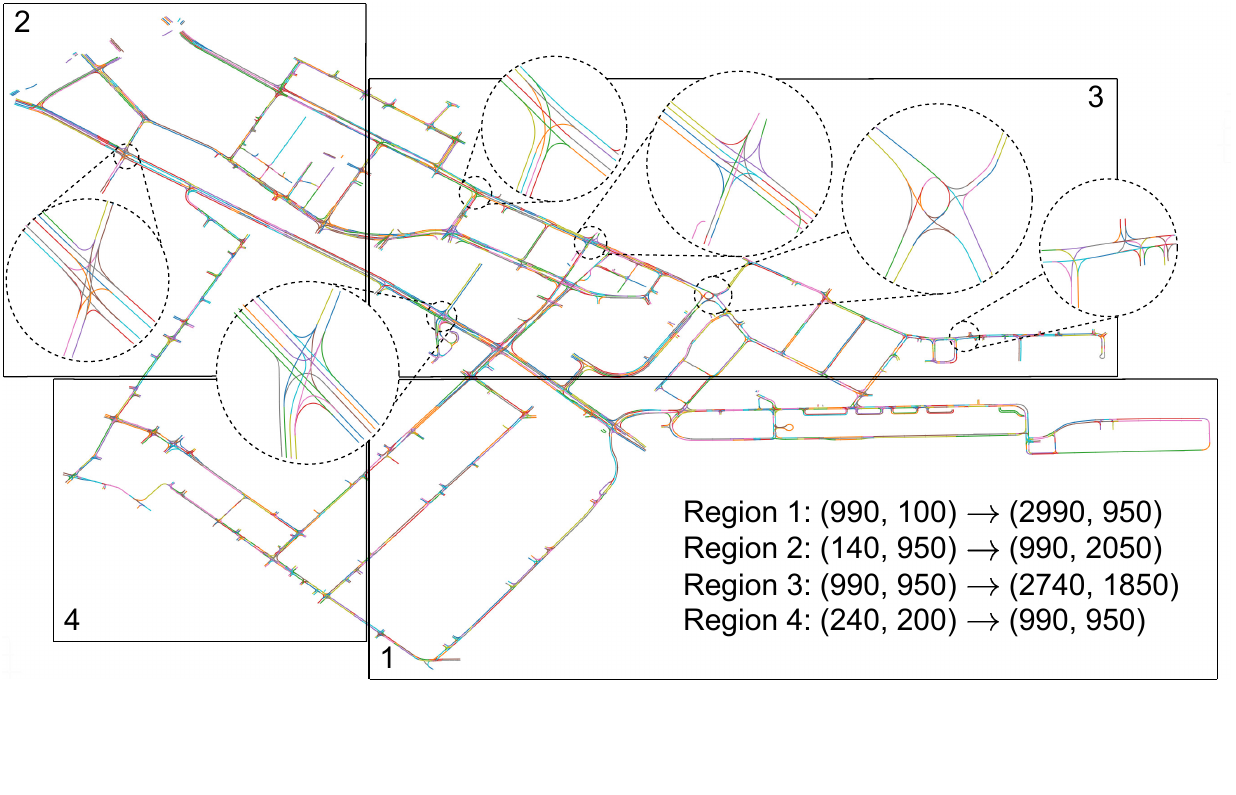}
\caption{Samples are partitioned into four nonoverlapping regions. Regions are specified by bottom-left and top-right corners in world coordinates.}
\label{fig:nuscenes_regions}
\vspace{-6mm}
\end{figure}

We evaluate the model performance on the right-side driving daytime Boston scenes in the nuScenes dataset~\cite{caesar2020nuscenes} similar to our baseline methods~\cite{can2021stsu, zurn2021lanegraphnet}.
The observation accumulation method described in Sec.~\ref{sec:partial_world_state_representation} generates a partially observed training sample $x$ every 1 m using accumulated observations from six 360$^{\circ}$ FoV RGB cameras and a top-mounted 32 beam lidar and a single pretrained semantic segmentation model \cite{karlsson2023pred_wm}. Each $x$ is augmented 20 times. Partitioning the generated training samples into the nonoverlapping regions shown in Fig.~\ref{fig:nuscenes_regions} results in 60,960 (34.7 \%), 40,960 (23.3 \%), and 73,780 (42.0 \%) samples for regions 1 to 3. Evaluation region 4 contains samples generated every 10 m without augmentation. We use a semantic segmentation model pretrained on two different public datasets~\cite{karlsson2023pred_wm}. We accumulate observations using ground truth pose information to reduce engineering effort, as prior work demonstrates the feasibility of accumulation based on pose estimation~\cite{karlsson2023pred_wm}. The plausible world state model input representation $\hat{x}$ consists of a five-layered 256$\times$256 grid map encompassing a 51.2$\times$51.2 m region similar to prior work~\cite{zurn2021lanegraphnet}. 

We conduct a model hyperparameter study and find that a smaller 1.4 M parameter model generalizes best. The model as depicted in Fig.~\ref{fig:dslp_model} has a common 8-layered CNN encoder with filter count increasing from 16 to 256, and two 8-layered CNN decoders with bilinear upsampling and filter count decreasing from 64 to 8.
See the code for further implementation details.

We use the following benchmarks to evaluate our DSLP model. We compare the global navigation pattern inference performance against the two most relevant and recently published SOTA supervised models STSU~\cite{can2021stsu} and LaneGraphNet~\cite{zurn2021lanegraphnet}. Both baselines are trained on nuScenes data~\cite{caesar2020nuscenes} to predict lane graphs using complete ground truth graphs as supervision. We compare the local probability field estimation performance against the prior self-supervised SOTA model called DSLA~\cite{karlsson2020dsla}.

\textbf{Local probability field estimation.}
We evaluate the predicted soft lane $\hat{Y}$ and directional $\hat{W}$ probability fields by computing the summed negative log-likelihood (NLL) of the ground truth lane map using (\ref{eq:nll_slp}) and (\ref{eq:nll_dp}). Lower NLL means the ground truth lane map is more likely according to the model. Directional accuracy measures the ratio of elements within $\pm 45^{\circ}$ of the ground truth direction.

\textbf{Global navigational pattern inference.}
We evaluate the usefulness of the predicted probability fields for inferring global navigational patterns by computing the intersection over union (IoU) and F1 score between the maximum likelihood graph and ground truth lane map.
Our method does not consider the spacing of graph nodes as an integral part of navigational patterns and thus does not view node displacement as a relevant performance metric.

\textbf{Ablation studies.}
We evaluate the advantage of our proposed predictive world modeling approach~\cite{karlsson2023pred_wm} for learning navigational patterns from sampled plausible completed worlds $\hat{x}$ instead of partially observed worlds $x$. We conduct an experiment using unaugmented samples to quantify the performance contribution of our geometric data augmentation method~\cite{karlsson2020dsla} on real-world data.
We conduct experiments on dataset splits including a different number of regions to estimate how performance increases with additional data.


\section{Results}

\begin{figure} 
\centering
\includegraphics[width=0.48\textwidth]{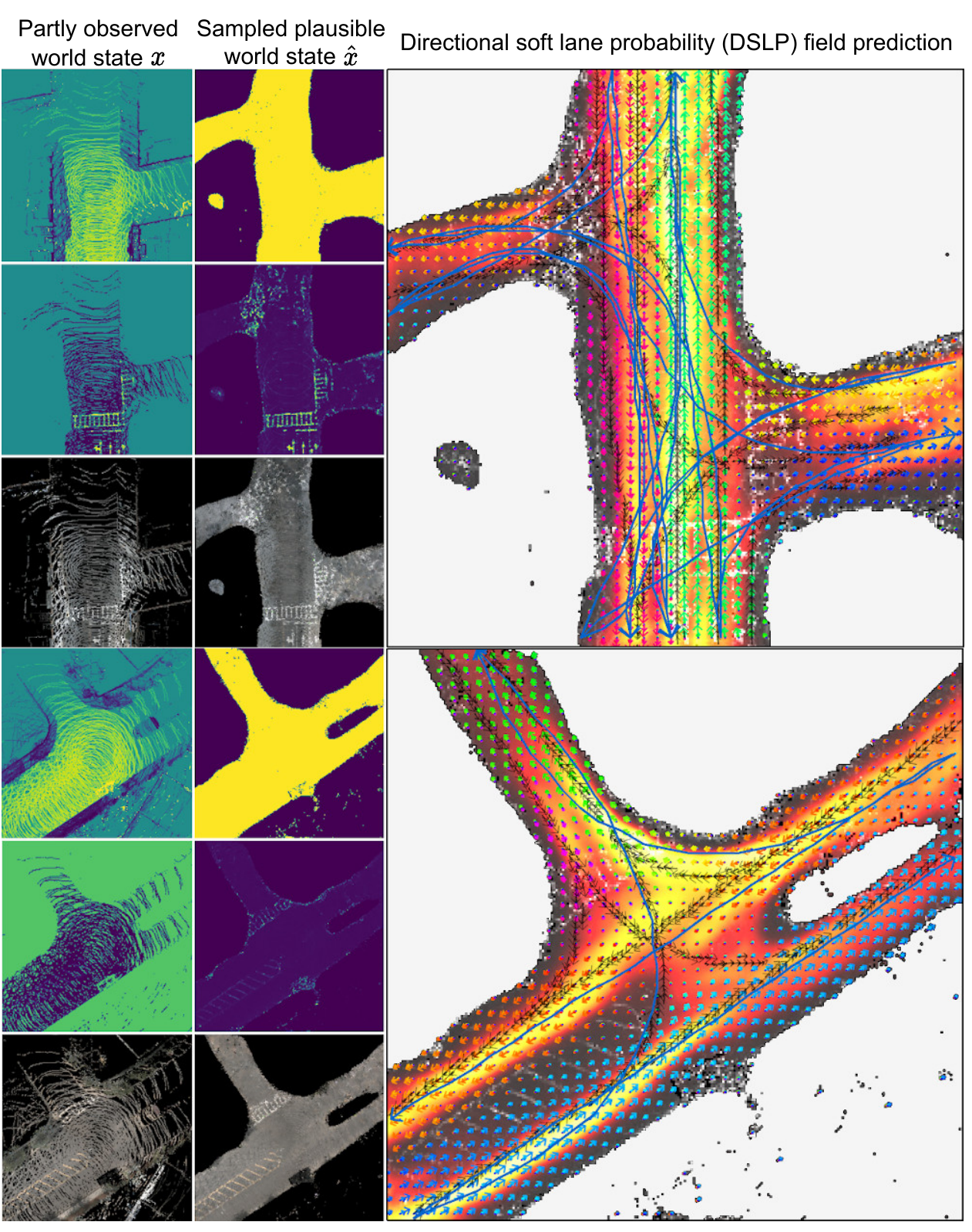}
\vspace{-8mm}
\caption{Model output visualizations. The left column shows accumulated partial observations $x$. The middle column shows plausible world states $\hat{x}$ sampled from $x$. The right column visualizes the predicted probability fields $\hat{Y}$ and $\hat{W}$ as well as the maximum likelihood graph.}
\label{fig:output_viz}
\vspace{-6mm}
\end{figure}

\textbf{Local probability field estimation.} Table~\ref{tab:dense_performance} presents evaluation results for the predicted probability fields. Our proposed DSLP model optimized with the information balance regularizer $\alpha_{IB}$ (\ref{eq:alpha_sla}) predicts the least biased probability field among all models trained and evaluated on accumulated past observation inputs. We conclude that the probabilistic objective (\ref{eq:slp_objective}) substantially reduces bias compared with the non-probabilistic DSLA affordance objective~\cite{karlsson2020dsla}.
Training and evaluating on accumulated past and future observation inputs in an offline map creation manner (i.e. full obs.) reduces bias, demonstrating that more comprehensively observed environments result in better performance. 
We performed experiments with different constant $\alpha$ values to demonstrate the merit of the proposed hyperparameter-free regularizer $\alpha_{IB}$ (\ref{eq:alpha_sla}). The best constant weight $\alpha$ value 0.1, found over five hyperparameter experiments, results in worse performance than using $\alpha_{IB}$.
We demonstrate the merit of dynamic, per-sample computed $\alpha_{IB}$ values (\ref{eq:alpha_sla}) by running an experiment with the constant mean $\alpha_{IB}$ value 0.122 computed over all training samples, which results in worse performance.
See Fig.~\ref{fig:output_viz} for probability field visualizations.



\begin{table}[t]
\caption{Performance of predicted local probability fields}
\vspace{-4mm}
\begin{center}
\begin{tabular}{|cc|c|c|c|c|}
    \hline
     \multicolumn{2}{|c|}{} & $\text{NLL}_{\text{SLP}}$ & $\text{NLL}_{\text{DP}}$ & NLL & Dir. acc. \\
    \hline
    \multicolumn{2}{|c|}{DSLA~\cite{karlsson2020dsla}} & 2.499 & 12.596 & 15.095 &  0.864 \\
    \hline
    \multirow{4}{*}{DSLP} & const $\alpha$ & 0.423 & 12.241 & 12.663 &  0.855 \\
    \cline{2-6}
    & mean $\alpha_{IB}$ & 0.444 & 12.038 & 12.482 &  0.881 \\
    \cline{2-6}
    & $\alpha_{IB}$ & 0.556 & 11.769 & 12.325 &  0.892 \\
    \cline{2-6}
    & full obs. & \textbf{0.539} & \textbf{11.666} & \textbf{12.205} & \textbf{0.900} \\
    \hline
\end{tabular}
\label{tab:dense_performance}
\end{center}
\vspace{-3mm}
\end{table}

\begin{table}[t]
\caption{Performance of global navigational pattern inference}
\vspace{-4mm}
\begin{center}
\begin{tabular}{|cc|c|c|}
    \hline
      \multicolumn{2}{|c|}{} & IoU & F1 score \\
    \hline
    \multicolumn{2}{|c|}{STSU~\cite{can2021stsu}} & 0.389 & 0.560 \\
    \hline
    \multicolumn{2}{|c|}{LaneGraphNet~\cite{zurn2021lanegraphnet}} & 0.420 & 0.574 \\
    \hline
    \multicolumn{2}{|c|}{DSLA~\cite{karlsson2020dsla}} & 0.427 (0.128) & 0.839 (0.07) \\
    \hline
     \multirow{4}{*}{DSLP} & constant $\alpha$ & 0.418 (0.146) & \textbf{0.853} (0.08) \\
    \cline{2-4}
    & mean $\alpha_{IB}$ & 0.410 (0.147) & 0.846 (0.08) \\
    \cline{2-4}
    & $\alpha_{IB}$ & 0.442 (0.125) & 0.834 (0.07) \\
    \cline{2-4}
    & full obs. & \textbf{0.454 (0.128)} & 0.839 (0.08) \\
    \hline
\end{tabular}
\label{tab:graph_performance}
\end{center}
\vspace{-7.5mm}
\end{table}

\textbf{Global navigational pattern inference.} Table~\ref{tab:graph_performance}
presents results showing that the maximum likelihood graph fitted to the probability field predicted by our self-supervised DSLP and prior DSLA model~\cite{karlsson2020dsla} from partially observed world representations $x$, outperforms the supervised SOTA baselines STSU~\cite{can2021stsu} and LaneGraphNet~\cite{zurn2021lanegraphnet} trained on ground truth lane graphs. 
Our self-supervised method not only improves upon the supervised baseline results while limited to the same training data domain, but is also a scaleable solution for real-world mobile robotics as the model can improve by continual learning from new observational experience. While the baselines do not specify train and evaluation regions for an ideal comparison, our experiments in Table~\ref{tab:data_amount} show our model surpassing the supervised baseline methods also when training on one region only, demonstrating that the exact train and evaluation region split is not critical for achieving our favorable results.
We note that the probabilistic DSLP model outperforms the non-probabilistic DSLA affordance model~\cite{karlsson2020dsla}, 
the proposed regularizer $\alpha_{IB}$~(\ref{eq:alpha_sla}) outperforms the best constant hyperparameter regularizer $\alpha$ and the mean $\alpha_{IB}$ value,
and that more comprehensively observed environments result in better performance. See Fig.~\ref{fig:output_viz} for inferred navigational path visualizations.

\textbf{Ablation studies.}
Table~\ref{tab:ablation_study} shows that leveraging the predictive world model (WM)~\cite{karlsson2023pred_wm} and 
proposed data augmentation (Aug.)~\cite{karlsson2020dsla} method reduces bias in the predicted probabilistic fields. We note that the unaugmented experiment generates output biased towards ego-agent trajectories, resulting in worse overall NLL while the maximum likelihood graph remains accurate. We believe this indicates the potential to further improve the graph generation algorithm to better leverage the more accurate probability field prediction. We do not explicitly evaluate the performance of the world model itself as this is done in prior work~\cite{karlsson2023pred_wm}.

Table~\ref{tab:data_amount} shows that increased observational experience reduces bias in the predicted probability field, providing evidence that the model can be trained to infer an unbiased probability prediction in the limit of infinite data


\textbf{Inference time.} We analyze the time taken for one iteration of our proposed system as follows. The mean inference time for the predictive world model and DSLP model is 0.175 sec and 0.017 sec, resulting in a total mean time of 0.192 sec per iteration or 5.21 Hz on an RTX 4090 GPU. We conclude that our method is feasible to run in real-time as it introduces a 0.192 sec overhead with a real-time SLAM implementation~\cite{vizzo2022kiss-icp} operating faster than sensor frame rates.

\section{Conclusion}

In this paper, we present the first SSL method for training a model to infer navigational patterns in real-world environments from partial observations while achieving better performance than SOTA supervised baselines.
Here we identify limitations and directions for future work.
The representation of spatially small but semantically important environmental cues, such as road markings, is inefficiently represented by uniform grid maps. Traffic information on signs is not represented at all. We propose to instead detect and semantically draw road markings and signs in the input representation.
Graph generation can be improved by inferring start and end points within the BEV, sampling higher-order splines, and decomposing splines into a sparse graph~\cite{karlsson2021gdsla}.
Understanding navigational patterns may require a temporal memory of past observations to resolve ambiguity. We propose an additional module that maintains a latent environment encoding by learning from sequences instead of i.i.d. data.


\begin{table}[t]
\caption{Ablation studies}
\vspace{-4mm}
\begin{center}
\begin{tabular}{|c|c|c|c|c|c|c|}
    \hline
     WM & Aug. & $\text{NLL}_{\text{SLP}}^{\ast}$ & $\text{NLL}_{\text{DP}}$ & $\text{NLL}^{\ast}$ & Dir. acc. & IoU \\ 
    \hline
    \cmark & \cmark & 0.189 & \textbf{11.769} & \textbf{11.958} & \textbf{0.892} & 0.442 \\ 
    \hline
    \xmark & \cmark & 0.266 & 12.785 & 13.051 & 0.853 & 0.223 \\ 
    \hline
    \cmark & \xmark & \textbf{0.167} & 13.764 & 13.931 & 0.848 & \textbf{0.453} \\ 
    \hline
   \multicolumn{6}{l}{$^{\ast}$Mean over all elements}
\end{tabular}
\label{tab:ablation_study}
\end{center}
\vspace{-3mm}
\end{table}

\begin{table}[t]
\caption{Performance with varying data amounts}
\vspace{-4mm}
\begin{center}
\begin{tabular}{|c|c|c|c|c|c|}
    \hline
     \# Regions & $\text{NLL}_{\text{SLP}}$ & $\text{NLL}_{\text{DP}}$ & NLL & Dir. acc. & IoU \\ 
    \hline
    \{1\} & \textbf{0.478} & 12.696 & 13.174 & 0.861 & 0.423 \\ 
    \hline
    \{1, 2\} & 0.544 & 12.013 & 12.557 & 0.874 & \textbf{0.444} \\ 
    \hline
    \{1, 2, 3\} & 0.556 & \textbf{11.769} & \textbf{12.325} & \textbf{0.892} & 0.442 \\ 
    \hline
\end{tabular}
\label{tab:data_amount}
\end{center}
\vspace{-7.5mm}
\end{table}





\section*{Acknowledgment}

This work was financially supported by JST SPRING, Grant Number JPMJSP2125. The authors would like to take this opportunity to thank the ``Interdisciplinary Frontier Next-Generation Researcher Program of the Tokai Higher Education and Research System''.


\begin{thebibliography}{99}

\bibitem{krause2013robot_navigation} T. Krause, A. Pandey, R. Alami, and A. Kirsch, Human-aware robot navigation: A survey, Robotics and Autonomous Systems 61(12), 1726-1743, 2013.
%
\bibitem{sheif2016hdmaps} H. Sheif and X. Hu, Autonomous Driving in the iCity - HD Maps as a Key Challenge of the Automotive Industry, Engineering 2, 2016.
%
\bibitem{bojarski2016e2e_self_driving} M. Bojarski, et al., End to End Learning for Self-Driving Cars, arXiv preprint, 2016.
%
\bibitem{amin2018variational_end2end} A. Amini, G. Rosman, S. Karaman, and D. Rus, Variational End-to-End Navigation and Localization, ICRA, 2018.
%
\bibitem{bansal2018} M. Bansal, A. Krizhevsky, and A. Ogale, ChauffeurNet: Learning to Drive by Imitating the Best and Synthesizing the Worst, arXiv, 2018.
%
\bibitem{zurn2021lanegraphnet} J. Z\"urn, J. Vertens, and W. Burgard, Lane Graph Estimation for Scene Understanding in Urban Driving, RA-L, 2021.
%
\bibitem{can2021stsu} Y. Can, A. Liniger, D. Paudel, and L. Van Gool, Structured Bird’s-Eye-View Traffic Scene Understanding from Onboard Images, ICCV, 2021.
%
\bibitem{gibson1979visual_system} J. Gibson, The ecological approach to visual perception, Houghton Mifflin, Boston, MA, 1979.
%
\bibitem{milner2008visual_systems} D. Milner and M. Goodale, Two visual systems re-viewed, Neuropsychologia 46, 3, pp. 774-785, 2008.
%
\bibitem{han2022modeling_ventral_dorsal} Z. Han and A. Sereno, Modeling the Ventral and Dorsal Cortical Visual Pathways Using Artificial Neural Networks, Neural Computation 34, 1, pp. 138–171, 2022.
%
\bibitem{milner2012visual_processing_conscious} D. Milner, Is visual processing in the dorsal stream accessible to consciousness?, Proc Biol Sci, 279, pp. 2289-2298, 2012.
%
\bibitem{sheth2016visual_pathways_sampling_space} B. Sheth and R. Young, Two Visual Pathways in Primates Based on Sampling of Space: Exploitation and Exploration of Visual Information, Front. Integr. Neurosci. 10, 2016.
%
\bibitem{lesort2020cont_learn_robot} T. Lesort, V. Lomonaco, A. Stoian, D. Maltoni, and D. Filliat, Continual learning for robotics: Definition, framework, learning strategies, opportunities and challenges, Information fusion, vol. 58, 2020.
%
\bibitem{minh2015dqn} V. Mnih, K. Kavukcuoglu, D. Silver, A. Rusu, J. Veness, M. Bellemare, et al., Human-level control through deep reinforcement learning, Nature, vol. 518, pp. 529--533, 2015.
%
\bibitem{karlsson2023pred_wm} R. Karlsson, A. Carballo, K. Fujii, K. Ohtani, and K. Takeda, Predictive World Models from Real-World Partial Observations, MOST, 2023.
%
\bibitem{karlsson2020dsla} R. Karlsson and E. Sjoberg, Learning a Directional Soft Lane Affordance Model for Road Scenes Using Self-Supervision, IV, 2020.
%
\bibitem{salzmann2019online_path_generation} T. Salzmann, J. Thomas, T. K\"uhbeck, J. Sung, S. Wagner, and A. Knoll, Online Path Generation from Sensor Data for Highly Automated Driving Functions, ITSC, 2019.
%
\bibitem{baumann2018ego_path} U. Baumann, C. Guiser, M. Herman, and J. Zollner, Predicting ego-vehicle paths from environmental observations with a deep neural network, ICRA, 2018.
%
\bibitem{barnes2017ws_path_proposal} D. Barnes, W. Maddern, and I. Posner, Find your own way: Weakly supervised segmentation of path proposals for urban autonomy, ICRA, 2017.
%
\bibitem{ort2020maplite} T. Ort, et al., MapLite: Autonomous Intersection Navigation Without a Detailed Prior Map, RL-L, 2020.
%
\bibitem{casas2021mp3} S. Casas, A. Sadat, and R. Urtasun, MP3: A Unified Model to Map, Perceive, Predict, and Plan, CVPR, 2021.
%
\bibitem{prez-higueras2018point_to_point_path} N. Prez-Higueras, F. Caballero, and L. Merino, Learning human-aware path planning with fully convolutional networks, ICRA, 2018.
%
\bibitem{karaman2011RRT_star} S. Karaman, and E. Frazzoli, Sampling-based algorithms for optimal motion planning, The Int. J. of Robotics Research, 30(7), 2011.
%
\bibitem{kitani2012activity_forcasting} K. Kitani, B. Ziebart, A. Bagnell, and M. Hebert, Activity forecasting, ECCV, 2012.
%
%
\bibitem{ratliff2009learch} N. Ratliff, D. Silver, and J. Bagnell, Learning to search: Functional gradient techniques for imitation learning, Autonomous Robots 27(1),
25-53, 2009.
%
\bibitem{homayounfar2018hran} N. Homayounfar, W. Ma, S. Lakshmikanth, and R. Urtasun, Hierarchical Recurrent Attention Networks for Structured Online Maps, CVPR, 2018.
%
\bibitem{homayounfar2019dagmapper} N. Homayounfar, J. Liang, W. Ma, J. Fan, X. Wu, and R. Urtasun, DAGMapper: Learning to Map by Discovering Lane Topology, ICCV, 2019.
%
\bibitem{guo2020genlanenet} Y. Guo et al., Gen-LaneNet: A Generalized and Scalable Approach for 3D Lane Detection, ECCV, 2020.
%
\bibitem{zhang2021hierarchical_road_topology_learning} L. Zhang, et al., Hierarchical Road Topology Learning for Urban Map-less Driving, IROS, 2022.
%
\bibitem{mi2021hdmapgen} L. Mi et al., HDMapGen: A Hierarchical Graph Generative Model of High Definition Maps, CVPR, 2021.
%
\bibitem{karlsson2021gdsla} R. Karlsson, D. Wong, S. Thompson, and K. Takeda, Learning a Model for Inferring a Spatial Road Lane Network Graph using Self-Supervision, ITSC, 2021.
%
\bibitem{li2022hdmapnet} Q. Li, Yu. Wang, Yi. Wang, and H. Zhao, HDMapNet: An Online HD Map Construction and Evaluation Framework, ICRA, 2022.
%
\bibitem{liu2023vectormapnet} Y. Liu, Y. Yuan, Yu., Wang, Yi. Wang, and H. Zhao, VectorMapNet: End-to-end Vectorized HD Map Learning, arXiv preprint, 2023.
%
\bibitem{liaomaptr} B. Liao, et al., MapTR: Structured Modeling and Learning for Online Vectorized HD Map Construction, ICRL, 2023.
%
\bibitem{shin2023instagram} J. Shin, F. Rameau, H. Jeong, and D. Kum, InstaGraM: Instance-level Graph Modeling for Vectorized HD Map Learning, arXiv preprint, 2023.
%
%
\bibitem{ort2022maplite2} T. Ort, J. Walls, S. Parkinson, I. Gilitschenski, and D. Rus, MapLite 2.0: Online HD Map Inference Using a Prior SD Map, RA-L, 2022.
%
\bibitem{pomerleau1988alvinn} D. Pomerleau, ALVINN: An Autonomous Land Vehicle in a Neural Network, NIPS, 1988.
%
\bibitem{kendall2019end2end_deeprl} A. Kendall, et al., Learning to Drive in a Day, ICRA, 2019.
%
\bibitem{henaff2019model_predictive_policy_learning} M. Henaff, A. Canziani, and Y. LeCun, Model-Predictive Policy Learning with Uncertainty Regularization for Driving in Dense Traffic, ICLR, 2019.
%
\bibitem{chen2021world_on_rails} D. Chen, V. Koltun, and P. Kr\"ahenb\"uhl, Learning to drive from a world on rails, ICCV, 2021.
%
\bibitem{mar1982human_hierarchical_vision_repr} D. Marr, Vision: A computational investigation into the human representation and processing of visual information, W.H. Freeman, 1982.
%
\bibitem{besl1992icp} P. J. Besl and N. D. McKay, A method for registration of 3-D shapes, IEEE Transactions on Pattern Analysis and Machine Intelligence, vol. 14, no. 2, pp. 239-256, 1992.
%
\bibitem{vizzo2022kiss-icp} I. Vizzo, T. Guadagnino, B. Mersch, L. Wiesmann, J. Behley, and C. Stachniss, KISS-ICP: In Defense of Point-to-Point ICP, RA-L, 2022.
%
\bibitem{smith1986slam} R. Smith and P. Cheeseman, On the Representation and Estimation of Spatial Uncertainty, The Int. J. of Robotics Research, 5(4), 1986.
%
%
\bibitem{child2021vdvae} R. Child, Very Deep VAEs Generalize Autoregressive Models and Can Outperform Them on Images, ICLR, 2021.
%
\bibitem{kingma2013vae} D. Kingma and M. Welling, Auto-Encoding Variational Bayes, CoRR, 2013.
%
\bibitem{marino2019pred_coding_vae_bio} J. Marino, Predictive Coding, Variational Autoencoders, and Biological Connections, Neural Computation, vol. 34, pp. 1--44, 2019.
%
\bibitem{ronneberger2015unet} O. Ronneberger, P. Fischer, and T. Brox, U-net: Convolutional networks for biomedical image segmentation, MICCAI, 2015.
%
\bibitem{murphy2022pml1} K. Murphy, Probabilistic Machine Learning: An Introduction, MIT Press, 2022.
%
%
%
\bibitem{caesar2020nuscenes} H. Caesar, et al., nuScenes: A multimodal dataset for autonomous driving, CVPR, 2020.
\end{thebibliography}
\end{document}